\documentclass[11pt]{article}

\usepackage[margin=1in]{geometry}
\usepackage[T1]{fontenc}
\usepackage{newtxtext,newtxmath}
\usepackage{amsmath, bm}
\usepackage{booktabs}
\usepackage{longtable}
\usepackage{array}
\usepackage{adjustbox}
\usepackage{algorithm}
\usepackage{algpseudocode}
\usepackage{hyperref}
\usepackage[numbers]{natbib}
\usepackage{makecell}
\usepackage{microtype}
\usepackage{setspace}
\usepackage{titlesec}
\usepackage{caption}
\usepackage{enumitem}
\usepackage{xcolor}
\usepackage{placeins}

\definecolor{headingblue}{RGB}{22,52,92}
\setlist[itemize]{topsep=0.35em,itemsep=0.18em,parsep=0pt,leftmargin=1.5em}
\titleformat{\section}{\large\bfseries\color{headingblue}}{\thesection}{0.7em}{}
\titleformat{\subsection}{\normalsize\bfseries\color{headingblue}}{\thesubsection}{0.65em}{}
\titleformat{\subsubsection}{\normalsize\itshape\bfseries}{\thesubsubsection}{0.6em}{}
\titlespacing*{\section}{0pt}{1.6em plus 0.25em minus 0.15em}{0.65em}
\titlespacing*{\subsection}{0pt}{1.1em plus 0.2em minus 0.1em}{0.45em}
\titlespacing*{\subsubsection}{0pt}{0.9em plus 0.15em minus 0.1em}{0.35em}

\title{Generalized Gibbs Ensemble Weighting for Forecast Combination}

\author{
Prasen R. Nuthanakaluva\\
Independent Researcher\\
Oxford, United Kingdom
\and
Nava K. Gaddam\\
Department of Physics\\
Utrecht University\\
Utrecht, Netherlands
}

\date{Preprint}

\begin{document}

\maketitle

\begin{abstract}
Forecast combination is a reliable way to improve predictive performance when several forecasting models are available. Simple aggregation rules such as the mean, median, trimmed mean, inverse-loss weighting, and exponential weighting are often strong baselines, but their relative performance can vary across datasets, forecast horizons, deployment settings, and levels of disagreement among base forecasters.

We develop Generalized Gibbs Ensemble Weighting (GGEW), a probabilistic framework that treats forecasting models as experts and assigns ensemble weights using a Gibbs-style exponential transformation of normalized predictive loss. The framework extends this basic weighting rule through numerical stabilization, diversity-aware score corrections, and online hyperparameter adaptation.

GGEW produces a family of related methods, including Stable Gibbs weighting, Directional Gibbs-NCL, and Symmetric Gibbs-NCL. These variants share one core algorithm and differ only in the score used inside the exponential weighting rule. For sequential deployment, we adopt a UCB-style bandit mechanism, called online Local-UCB, to adapt the learning rate, diversity strength, and Gibbs variant without evaluating the full hyperparameter grid at every prediction step.

We evaluate GGEW on official M4 competition forecast submissions and external rolling-origin deployment experiments using Monash Traffic Hourly, Electricity Hourly, and Solar Weekly datasets. Results suggest that Gibbs-style adaptive weighting is a useful and competitive tool across several benchmark settings, although its relative performance varies across datasets, forecast horizons, deployment protocols, and forecast disagreement groups. The contribution is not a universal dominance claim, but a framework and empirical study motivating further investigation of when adaptive Gibbs-style forecast combination is useful.
\end{abstract}

\section{Introduction}

Forecast combination is a central problem in machine learning and time series forecasting. A single forecasting model rarely captures every useful structure in a dataset. Different models may respond differently to trend, seasonality, intermittency, noise, structural breaks, and local scale changes. Combining forecasts can therefore reduce variance, improve robustness, and exploit complementary predictive behavior across models.

Classical forecast-combination methods include equal-weight averaging, median aggregation, trimmed means, inverse-loss weighting, exponential weighting, and historical best-model selection. These methods are simple and often difficult to beat. However, they typically apply a fixed aggregation rule across all forecasting situations. This is restrictive because real forecasting environments differ substantially in the degree and structure of cross-model disagreement.

In some cases, models agree closely and only small differences in historical predictive performance separate them. In other cases, model forecasts diverge strongly, and a few predictions may be extreme relative to the rest of the forecast pool. These situations should not necessarily be handled by the same aggregation rule. A robust median may be preferable when forecasts are highly dispersed, while adaptive probabilistic weighting may be useful when disagreement is moderate and structured.

This paper proposes Generalized Gibbs Ensemble Weighting as a framework for adaptive forecast combination. The method formulates Gibbs-style exponential allocation as an online expert-weighting procedure for forecast aggregation. Predictive loss is treated as a score, lower scores receive higher probability mass, and the resulting weights form a convex ensemble.

The main contributions are:
\begin{itemize}
    \item A unified Gibbs-style probabilistic weighting formulation for forecast combination.
    \item A stable exponentiated-gradient update that preserves the probability simplex without projection.
    \item A diversity-aware extension that incorporates negative-correlation-style corrections into the weighting score.
    \item A symmetric diversity variant designed to reward useful forecast non-redundancy in a balanced way.
    \item An online Local-UCB procedure for adapting hyperparameter states during sequential forecasting.
    \item A regime-based empirical protocol that evaluates performance under low, medium, and high disagreement forecast regimes.
\end{itemize}

The main empirical question is whether adaptive Gibbs-style weighting can improve forecast combination under different levels of cross-model disagreement and in sequential deployment settings. The paper deliberately avoids claiming that one method dominates universally. Instead, it studies how adaptive weighting behaves across datasets, horizons, and disagreement structures.

\section{Related Work}

\subsection{Forecast Combination}

Forecast combination has a long history in forecasting research \citep{bates1969combination}. Simple averages are strong because they reduce variance and avoid overfitting the combination rule. Median and trimmed-mean ensembles provide robustness against extreme or unstable forecasts. Performance-based weighting methods use historical losses to allocate more weight to models that have recently performed well.

The proposed method belongs to the broad class of adaptive forecast-combination methods, but differs from standard inverse-loss weighting and simple exponential weighting by using a stabilized Gibbs-style score, an iterative simplex-preserving update, and optional diversity corrections.

\subsection{Online Expert Aggregation}

Online expert aggregation studies sequential prediction with multiple experts. Methods such as exponential weights, Hedge, multiplicative weights, and mirror descent update expert weights after observing losses \citep{vovk1990aggregating, cesa2006prediction}. These methods motivate the use of exponentiated updates, since they naturally preserve non-negativity and sum-to-one constraints.

GGEW uses this same optimization geometry. The update is performed in log-weight space and then normalized, which keeps weights on the simplex and avoids the instability of additive updates followed by ad hoc rescaling.

\subsection{Diversity-Aware Ensembles and Negative Correlation Learning}

Ensemble accuracy depends not only on the accuracy of each base model, but also on the dependence structure among model errors. Negative correlation learning encourages ensemble members to behave differently when diversity is useful \citep{liu1999negative}. In forecast combination, this suggests that the weighting score should not depend only on individual squared loss, but may also include a term measuring how a forecast differs from the ensemble consensus.

This paper uses this idea to define Directional Gibbs-NCL and Symmetric Gibbs-NCL. These variants modify the loss score before applying the Gibbs-style exponential weighting rule.

\subsection{Gibbs-Style Exponential Allocation in Machine Learning}

The term Gibbs is used here in the standard machine-learning sense of exponential allocation over scores or losses. Given a set of scores, the Gibbs rule converts them into a probability distribution by exponentiating negative scores and normalizing. In this paper, the score is a stabilized predictive-loss quantity, optionally augmented by diversity terms.

\section{Problem Formulation}

Consider a forecasting problem with $M$ candidate forecasting models. At time $t$, model $m$ produces prediction
\begin{equation}
    \hat{y}_{t,m}.
\end{equation}

The ensemble forecast is a convex combination:
\begin{equation}
    \hat{y}_{t}^{\,ens} = \sum_{m=1}^{M} w_m \hat{y}_{t,m},
\end{equation}
where
\begin{equation}
    w_m \geq 0,
    \qquad
    \sum_{m=1}^{M} w_m = 1.
\end{equation}

Let $\widehat{Y} \in \mathbb{R}^{T \times M}$ denote a historical forecast matrix, where $\hat{y}_{t,m}$ is the prediction made by expert $m$ for historical observation $t$. Let $y_t$ denote the corresponding realized value. The task is to estimate ensemble weights from historical forecast errors and then apply those weights to future forecasts.

\section{Generalized Gibbs Ensemble Weighting}

\subsection{Gibbs-Style Allocation Rule}

Given model-specific scores $S_m$, the Gibbs-style allocation rule is
\begin{equation}
    w_m = \frac{\exp(-\eta S_m)}
    {\sum_{j=1}^{M} \exp(-\eta S_j)},
\end{equation}
where $\eta > 0$ controls the concentration of the allocation. Larger $\eta$ gives sharper emphasis to lower-score models, while smaller $\eta$ produces smoother weights across the ensemble.

Here, $S_m$ denotes the aggregate score assigned to base expert $m$ using the historical information available before the current forecast is made. In the simplest case, $S_m$ is the cumulative normalized predictive loss of expert $m$ over the available history.

For the generalized variants, let $v \in \{\mathrm{SG},\mathrm{DNCL},\mathrm{SNCL}\}$ index the variant type. Here $\mathrm{SG}$ denotes Stable Gibbs with no diversity correction, $\mathrm{DNCL}$ denotes directional Gibbs-NCL, and $\mathrm{SNCL}$ denotes symmetric Gibbs-NCL. The aggregate score is then obtained by summing the variant-specific score $G_{t,m}^{(v)}$ over the historical cases:
\begin{equation}
    S_m^{(v)}
    =
    \sum_{t=1}^{T} G_{t,m}^{(v)}.
\end{equation}
Thus, lower values of $S_m$ indicate historically better or more favorable experts under the chosen Gibbs variant, and these experts receive larger Gibbs weights.

The normalization factor is
\begin{equation}
    Z = \sum_{j=1}^{M} \exp(-\eta S_j),
\end{equation}
so that
\begin{equation}
    w_m = \frac{\exp(-\eta S_m)}{Z}.
\end{equation}

This guarantees
\begin{equation}
    w_m > 0,
    \qquad
    \sum_{m=1}^{M} w_m = 1.
\end{equation}

The ensemble forecast is then
\begin{equation}
    \hat{y}^{\,ens} = \sum_{m=1}^{M} w_m \hat{y}_m.
\end{equation}

\subsection{Predictive Loss Score}

For historical observation $t$, define the quadratic predictive loss
\begin{equation}
    L_{t,m} = \frac{1}{2}(\hat{y}_{t,m} - y_t)^2.
\end{equation}

Raw squared losses can be highly scale-dependent. Therefore, the loss is normalized within each historical row:
\begin{equation}
    \tilde{L}_{t,m} =
    \frac{L_{t,m}}
    {\max_{j} |L_{t,j}| + \varepsilon},
\end{equation}
where $\varepsilon > 0$ is a numerical stabilizer. Row-wise normalization prevents very large squared errors from dominating the exponential update and makes the algorithm more stable across heterogeneous time series.

\subsection{Generalized Score}

The generalized score used by the method is
\begin{equation}
    G_{t,m}^{(v)}
    =
    \tilde{L}_{t,m}
    +
    \lambda C_{t,m}^{(v)},
\end{equation}
where $v$ denotes the variant type, $C_{t,m}^{(v)}$ is a variant-specific correction term, and $\lambda \geq 0$ controls the strength of the correction.

The three variants are:
\begin{align}
    C_{t,m}^{\mathrm{SG}} &= 0,\\
    C_{t,m}^{\mathrm{DNCL}} &= \bar{\hat{y}}_t - \hat{y}_{t,m},\\
    C_{t,m}^{\mathrm{SNCL}} &= - (\hat{y}_{t,m} - \bar{\hat{y}}_t)^2,
\end{align}
where
\begin{equation}
    \bar{\hat{y}}_t = \frac{1}{M}\sum_{m=1}^{M} \hat{y}_{t,m}.
\end{equation}

The directional term is signed and can reward forecasts on one side of the ensemble mean. The symmetric term rewards non-redundancy through squared deviation from the ensemble mean. Since lower scores receive larger Gibbs allocation, the negative sign in the symmetric term lowers the score of forecasts that are distinct from the ensemble consensus.

The resulting generalized Gibbs allocation is
\begin{equation}
    w_m \propto \exp\left(
    -\eta \sum_{t=1}^{T} G_{t,m}^{(v)}
    \right).
\end{equation}

This single equation covers Stable Gibbs, Directional Gibbs-NCL, and Symmetric Gibbs-NCL.

\section{Weight Optimization and Gradient Structure}

The algorithm maintains a persistent vector of ensemble weights $\bm{w}$ on the simplex. The score matrix is used to update this vector iteratively.

Define
\begin{equation}
    A_{t,m} = \exp(-w_m G_{t,m}),
\end{equation}
and
\begin{equation}
    S_t = \sum_{j=1}^{M} A_{t,j}.
\end{equation}

A natural log-normalized objective is
\begin{equation}
    \mathcal{J}(\bm{w})
    =
    \sum_{t=1}^{T} \log \left(
    \sum_{j=1}^{M} \exp(-w_j G_{t,j})
    \right).
\end{equation}

The implemented algorithm uses a leave-one-out relative gradient signal. Let $g_m$ denote the relative gradient signal used to update the weight of expert $m$:
\begin{equation}
    g_m =
    \sum_{t=1}^{T}
    \frac{
    G_{t,m}(S_t - A_{t,m})
    }
    {
    (M-1)S_t + \varepsilon
    }.
\end{equation}

This compares each model against the remaining ensemble mass rather than only its own normalized exponential allocation. It is a relative performance signal: a model receives a larger update when its score is high relative to the rest of the ensemble.

Before updating the weights, the gradient is centered and clipped:
\begin{equation}
    g_m \leftarrow g_m - \frac{1}{M}\sum_{j=1}^{M}g_j,
\end{equation}
and
\begin{equation}
    g_m \leftarrow
    \min\{g_{clip}, \max\{-g_{clip}, g_m\}\}.
\end{equation}

\section{Stable Exponentiated-Gradient Update}

Let $k$ index the inner Gibbs optimization iteration. Each call to the Gibbs weighting routine runs $K$ such iterations; in the experiments $K=30$.

The weights are updated in log space:
\begin{equation}
    \log w_m^{(k+1)}
    =
    \log(w_m^{(k)}+\varepsilon)
    -
    \eta g_m.
\end{equation}

For numerical stability:
\begin{equation}
    \ell_m \leftarrow \ell_m - \max_j \ell_j,
\end{equation}
where $\ell_m = \log w_m^{(k+1)}$.

The weights are then normalized:
\begin{equation}
    w_m^{(k+1)} =
    \frac{\exp(\ell_m)}
    {\sum_{j=1}^{M}\exp(\ell_j)}.
\end{equation}

Equivalently,
\begin{equation}
    w_m^{(k+1)}
    \propto
    w_m^{(k)}\exp(-\eta g_m).
\end{equation}

This update guarantees
\begin{equation}
    w_m^{(k+1)} > 0,
    \qquad
    \sum_{m=1}^{M} w_m^{(k+1)} = 1.
\end{equation}

A small weight floor can optionally be applied after normalization:
\begin{equation}
    w_m \leftarrow (1 - M\delta)w_m + \delta,
\end{equation}
followed by renormalization.

\section{Algorithmic Recipe}

\begin{algorithm}[ht]
\caption{Generalized Gibbs Ensemble Weighting}
\begin{algorithmic}[1]
\Require Historical forecasts $\widehat{Y}$, historical outcomes $\bm{y}$, future forecasts $F$, variant $v$, learning rate $\eta$, correction strength $\lambda$, iterations $K$, stabilizer $\varepsilon$, clipping threshold $g_{clip}$, optional weight floor $\delta$
\Ensure Ensemble forecast $\hat{\bm{y}}^{\,ens}$
\State Initialize $w_m = 1/M$ for all models $m$
\For{$k = 1,\ldots,K$}
    \State Compute squared predictive loss $L_{t,m} = \frac{1}{2}(\hat{y}_{t,m}-y_t)^2$
    \State Compute the variant correction $C_{t,m}^{(v)}$
    \State Form generalized score $G_{t,m}^{(v)} = L_{t,m} + \lambda C_{t,m}^{(v)}$
    \State Normalize each historical row by $\max_m |G_{t,m}^{(v)}|+\varepsilon$
    \State Compute $A_{t,m} = \exp(-w_m G_{t,m}^{(v)})$ and $S_t = \sum_j A_{t,j}$
    \State Compute leave-one-out gradient
    \[
    g_m = \sum_t
    \frac{G_{t,m}^{(v)}(S_t-A_{t,m})}
    {(M-1)S_t+\varepsilon}
    \]
    \State Center and clip the gradient
    \State Update weights using exponentiated-gradient normalization
    \State Apply optional weight flooring and renormalize
\EndFor
\State \Return $\hat{\bm{y}}^{\,ens} = F\bm{w}$
\end{algorithmic}
\end{algorithm}

This pseudocode is intentionally written as a single algorithm because Stable Gibbs, Directional Gibbs-NCL, and Symmetric Gibbs-NCL are not separate algorithms; they are variants of the same weighting mechanism with different score corrections.

\section{Online Local-UCB Hyperparameter Adaptation}

The Gibbs family contains hyperparameters that may be difficult to set globally. Let
\begin{equation}
    a = (\eta, \lambda, v)
\end{equation}
denote a parameter state. Rather than selecting one fixed state using test-set performance, the online version adapts the active state sequentially as new outcomes are observed.

For each state $a$, maintain an exponentially smoothed loss estimate:
\begin{equation}
    \bar{\ell}_a(t)
    =
    \beta \bar{\ell}_a(t-1)
    +
    (1-\beta)\ell_a(t),
\end{equation}
where $\beta$ controls memory. Also maintain the number of times $n_a(t)$ that the state has been evaluated.

A lower-confidence loss score is computed as
\begin{equation}
    U_a(t)
    =
    \bar{\ell}_a(t)
    -
    c_{\mathrm{UCB}}\sqrt{
    \frac{\log(t+1)}
    {n_a(t)+1}
    }.
\end{equation}

Here $c_{\mathrm{UCB}}$ is the exploration constant. For a fixed time $t$, the values $\{U_a(t):a\in\mathcal{A}\}$ form a vector of uncertainty-adjusted loss scores over the candidate parameter states. Since the objective is to minimize loss, smaller values of $U_a(t)$ are preferred. The second term gives a temporary advantage to states that have been evaluated fewer times, encouraging exploration.

The tolerance parameter controls how close another state's uncertainty-adjusted score must be to the best score in order to be included in the local candidate set.

Before updating the smoothed losses, candidate state losses are scaled by the median loss among the evaluated candidate states at time $t$. This keeps the Local-UCB update numerically comparable across cases with different error scales.

The next state is selected by
\begin{equation}
    a_{t+1} = \arg\min_{a \in \mathcal{C}_t} U_a(t),
\end{equation}
where $\mathcal{C}_t$ is a local candidate set containing the current state, neighboring states, and any states close to the best uncertainty-adjusted score.

\begin{algorithm}[ht]
\caption{Online Local-UCB State Adaptation}
\begin{algorithmic}[1]
\Require Sequential evaluation cases, parameter state space $\mathcal{A}$, initial state $a_1$, memory $\beta$, exploration constant $c_{\mathrm{UCB}}$, tolerance $\tau$, minimum and maximum candidate budgets $K_{min},K_{max}$
\Ensure Online ensemble forecasts and state history
\State Initialize smoothed losses and visit counts for all states
\For{each forecasting case $t$}
    \State Construct a local candidate set around the active state
    \State Add additional states with promising uncertainty-adjusted loss scores
    \State Set the candidate budget between $K_{min}$ and $K_{max}$, using larger budgets early and smaller budgets later
    \State Evaluate Generalized Gibbs Ensemble Weighting only for selected candidate states
    \State Produce the forecast using the current active state
    \State When the realized outcome is observed, compute candidate losses
    \State Scale candidate losses by their median value
    \State Update smoothed loss estimates and visit counts
    \State Select the next active state using the Local-UCB rule
\EndFor
\State \Return forecasts, selected states, losses, and diagnostics
\end{algorithmic}
\end{algorithm}

\section{Computational Complexity}

Let $M$ be the number of base forecasting models, $T$ the number of historical observations, $K$ the number of Gibbs iterations, and $|\mathcal{C}_t|$ the number of candidate states evaluated online at time $t$.

The approximate cost per online forecasting case is
\begin{equation}
    O(|\mathcal{C}_t| K T M).
\end{equation}

A full-grid online method would cost
\begin{equation}
    O(|\mathcal{A}| K T M),
\end{equation}
where $|\mathcal{A}|$ is the total number of parameter states. Since the Local-UCB method is designed so that
\begin{equation}
    |\mathcal{C}_t| \ll |\mathcal{A}|,
\end{equation}
it is substantially cheaper than full-grid online adaptation while still allowing state changes when recent data support them.

\section{Disagreement-Regime Stratification}

Prior work has shown that diversity among forecasts is an important factor in forecast-combination performance \citep{kang2022forecast}. Motivated by this, the empirical protocol stratifies series by relative cross-model forecast disagreement.

For series $i$, horizon $h$, and model $m$, let $\hat{y}_{i,h,m}$ denote the forecast. The cross-model variance at horizon $h$ is
\begin{equation}
    V_{i,h}
    =
    \operatorname{Var}_m(\hat{y}_{i,h,m}).
\end{equation}

Because raw variance is scale-dependent, define the relative variance
\begin{equation}
    RV_{i,h}
    =
    \frac{
    \operatorname{Var}_m(\hat{y}_{i,h,m})
    }
    {
    \left(|\bar{y}_{i,h}|+\varepsilon\right)^2
    },
\end{equation}
where
\begin{equation}
    \bar{y}_{i,h}
    =
    \frac{1}{M}\sum_{m=1}^{M}\hat{y}_{i,h,m}.
\end{equation}

The series-level disagreement score is
\begin{equation}
    RV_i
    =
    \frac{1}{H}\sum_{h=1}^{H}RV_{i,h}.
\end{equation}

Series are then partitioned into high, medium, and low disagreement groups.

\section{Experimental Evaluation}

We evaluate Generalized Gibbs Ensemble Weighting in an online forecast-combination setting. Each method is treated as an online forecaster: ensemble weights are updated using only past observed cases, and Gibbs-family hyperparameters are adapted using the Local-UCB state-selection procedure.

The empirical evaluation uses two types of data. The first is based on official M4 competition forecast submissions \citep{makridakis2020m4}, where multiple independent submitted forecasts are available for each time series. The second uses external rolling-origin experiments on Monash Traffic Hourly, Electricity Hourly, and Solar Weekly datasets \citep{godahewa2021monash}. For the Monash datasets, ready-made multi-method forecast matrices were not available in the required forecast-matrix format, so we generated reproducible deterministic rolling baseline forecasts and then applied the ensemble methods to those forecasts.

\subsection{Compared Methods}

\begin{table}[ht]
\centering
\caption{Compared forecast-combination methods.}
\small
\setlength{\tabcolsep}{5pt}
\renewcommand{\arraystretch}{1.15}
\begin{adjustbox}{max width=\textwidth}
\begin{tabular}{p{0.30\linewidth}p{0.62\linewidth}}
\toprule
Category & Methods \\
\midrule
Simple baselines & mean, median, trimmed mean \\
Performance-based baselines & inverse-loss weighting, exponential weighting, best historical model \\
Selection baseline & top-3 average \\
Proposed online methods & Local-UCB Stable Gibbs, Local-UCB Stable Gibbs-NCL, Local-UCB Symmetric Gibbs \\
\bottomrule
\end{tabular}
\end{adjustbox}
\end{table}

The comparison methods are included as standard practical baselines rather than as new contributions. Let $\hat{y}_{h,m}$ denote the forecast from base expert $m$ at horizon $h$, and let $L_m$ denote the recent historical loss of expert $m$ computed from the available online history. The mean ensemble assigns equal weights to all base forecasts, while the median ensemble uses the horizon-wise median across experts. The trimmed mean removes the largest and smallest forecasts at each horizon before averaging the remaining forecasts.

The inverse-loss ensemble uses weights
\begin{equation}
    w_m = \frac{(L_m+\epsilon)^{-1}}{\sum_{j=1}^{M}(L_j+\epsilon)^{-1}},
\end{equation}
where $\epsilon$ is a small numerical constant. The exponential-weighted ensemble uses
\begin{equation}
    w_m = \frac{\exp(-\rho L_m)}{\sum_{j=1}^{M}\exp(-\rho L_j)},
\end{equation}
which follows the standard expert-weighting principle of assigning more weight to lower-loss experts. The best-historical-model baseline selects the expert with the smallest recent historical loss, and the top-3 average takes the simple average of the three experts with the smallest recent historical losses. These baselines cover equal weighting, robust aggregation, loss-based weighting, exponential expert weighting, and historical expert selection.

For the Local-UCB Gibbs-family methods, the candidate Gibbs states are selected online using fixed controller settings. Unless otherwise stated, we use $K_{\min}=3$, $c_{\mathrm{UCB}}=0.5$, $\beta=0.9$, tolerance $0.05$, and 30 inner Gibbs iterations. For Local-UCB Stable Gibbs, $K_{\max}=4$, reflecting the four learning-rate states without NCL correction. For Local-UCB Stable Gibbs-NCL and Local-UCB Symmetric Gibbs, $K_{\max}=10$, allowing a larger adaptive neighborhood over learning-rate and diversity-strength states. Candidate state losses are scaled by their median before updating the Local-UCB smoothed losses. The Gibbs numerical settings are fixed as gradient clipping threshold $g_{clip}=5$, weight floor $\delta=10^{-4}$, and numerical stabilizer $\varepsilon=10^{-12}$. These controller and numerical settings are fixed across all experiments and are not tuned separately for each dataset.

The candidate learning-rate values are $\eta \in \{0.01,0.05,0.1,0.2\}$. For the NCL variants, the candidate diversity strengths are $\lambda \in \{0.01,0.05,0.1,0.5\}$. Stable Gibbs uses no NCL correction, Stable Gibbs-NCL uses directional NCL, and Symmetric Gibbs uses the symmetric diversity correction.

\subsection{M4 Online Forecast-Matrix Experiments}

For M4, we use all valid official numbered point-forecast submissions containing the relevant frequency, giving 41 valid submitted forecasting systems after excluding submissions without usable rows for the target frequency. Each submitted forecast column is treated as a base forecasting expert. Since a submission may internally use different model classes or series-specific rules across different time series, an M4 expert should be interpreted as a submitted forecasting system rather than a single fixed parametric model. The corresponding M4 test actuals are used to evaluate the ensemble forecasts after each online case.

Series are stratified into high, medium, and low disagreement regimes using relative cross-model forecast variance. For each series and horizon, we compute the variance of forecasts across submitted systems and normalize it by squared forecast scale. The resulting horizon-level relative variances are averaged across the forecast horizon to obtain a single disagreement score per series. High disagreement series are selected from the top of this ranking, low disagreement series from the bottom, and medium disagreement series from the middle.

The online M4 stream is balanced across frequencies and regimes. For each frequency and disagreement regime, 4500 series are used, split into three independent blocks of 1500. Each block uses 500 warmup cases and 1000 evaluated cases. Therefore, each frequency/regime result is based on 3000 evaluated cases. For the M4 online stream experiments, Gibbs-family methods use a fixed historical memory budget of $T=200$ past observed cases when estimating ensemble weights. This is not treated as a tuned Gibbs hyperparameter; it is a computational memory budget used to make the online procedure scalable. The memory buffer is sampled only from previously observed cases, so no future information is used.

\begin{table}[ht]
\centering
\caption{M4 online forecast-matrix benchmark design.}
\begin{adjustbox}{max width=\textwidth}
\begin{tabular}{lllr}
\toprule
Dataset & Horizon & Forecast pool & Evaluated cases per regime \\
\midrule
M4 Yearly & 6 years & 41 submitted forecasting systems & 3000 \\
M4 Quarterly & 8 quarters & 41 submitted forecasting systems & 3000 \\
M4 Monthly & 18 months & 41 submitted forecasting systems & 3000 \\
\bottomrule
\end{tabular}
\end{adjustbox}
\end{table}

\subsection{Monash Rolling-Origin Deployment Experiments}

The Monash experiments test the online ensemble methods in rolling deployment simulations. Unlike M4, these datasets do not provide official multi-method submission matrices in the required format. We therefore generate deterministic rolling baseline forecasts and evaluate the ensemble methods on top of those forecasts.

\begin{table}[ht]
\centering
\caption{Monash rolling-origin deployment design.}
\small
\begin{adjustbox}{max width=\textwidth}
\begin{tabular}{llllrrr}
\toprule
Dataset & Series used & Initial history & Deployment period & Horizon & Evaluated cases & Base models \\
\midrule
Traffic Hourly & 100 sensors & 90 days & 270 days & 48 hours & 13,000 & 8 \\
Electricity Hourly & 100 clients & 90 days & 270 days & 24 hours & 26,500 & 8 \\
Solar Weekly & 137 series & 30 weeks & rolling weekly & 5 weeks & 1,781 & 9 \\
\bottomrule
\end{tabular}
\end{adjustbox}
\end{table}

For Traffic and Electricity, the deterministic base forecast pool contains last value, daily seasonal naive, weekly seasonal naive, recent 24-hour mean, recent 168-hour mean, 168-hour drift, daily profile, and weekly profile forecasts. These baselines are chosen because hourly traffic and electricity data contain strong daily and weekly structure. For Solar Weekly, nine deterministic rolling baselines are generated using the same principle: fast, reproducible forecasts that provide a diverse base pool without requiring expensive model refitting.

For the Monash rolling-origin experiments, the historical window is defined locally within each series using the previous five rolling forecast windows after an initial five-window warmup. The Monash experiments should therefore be interpreted as external online deployment tests of the ensemble framework, not as comparisons against official Monash benchmark forecast submissions.

\begin{table}[ht]
\centering
\caption{Deterministic rolling baseline forecast pool.}
\begin{tabular}{ll}
\toprule
Base forecast & Description \\
\midrule
last value & repeats most recent value \\
daily seasonal naive & repeats value from 24 hours ago \\
weekly seasonal naive & repeats value from 168 hours ago \\
24-hour mean & repeats recent 24-hour mean \\
168-hour mean & repeats recent weekly mean \\
168-hour drift & linear drift over recent week \\
daily profile & hour-of-day profile over recent 28 days \\
weekly profile & hour-of-week profile over recent 4 weeks \\
\bottomrule
\end{tabular}
\end{table}

\section{Results and Discussion}

\subsection{M4 Online Results}

\begin{table}[ht]
\centering
\caption{Top three M4 online methods by frequency and disagreement regime.}
\small
\setlength{\tabcolsep}{4pt}
\renewcommand{\arraystretch}{1.12}
\begin{adjustbox}{max width=\textwidth}
\begin{tabular}{p{0.24\linewidth}p{0.23\linewidth}p{0.23\linewidth}p{0.23\linewidth}}
\toprule
Dataset / Regime & Rank 1 & Rank 2 & Rank 3 \\
\midrule
M4 Yearly High & inverse-loss weighted & median & Local-UCB Stable Gibbs \\
M4 Yearly Medium & Local-UCB Symmetric Gibbs & Local-UCB Stable Gibbs & inverse-loss weighted \\
M4 Yearly Low & top-3 average & inverse-loss weighted & Local-UCB Stable Gibbs \\
M4 Quarterly High & median & trimmed mean & inverse-loss weighted \\
M4 Quarterly Medium & Local-UCB Symmetric Gibbs & Local-UCB Stable Gibbs-NCL & Local-UCB Stable Gibbs \\
M4 Quarterly Low & Local-UCB Stable Gibbs-NCL & Local-UCB Symmetric Gibbs & inverse-loss weighted \\
M4 Monthly High & Local-UCB Symmetric Gibbs & top-3 average & Local-UCB Stable Gibbs \\
M4 Monthly Medium & inverse-loss weighted & exponential weighted & trimmed mean \\
M4 Monthly Low & Local-UCB Stable Gibbs-NCL & inverse-loss weighted & Local-UCB Symmetric Gibbs \\
\bottomrule
\end{tabular}
\end{adjustbox}
\end{table}

The M4 results are mixed but encouraging. Gibbs-family methods win several regimes: Symmetric Gibbs wins M4 Yearly medium, M4 Quarterly medium, and M4 Monthly high; Stable Gibbs-NCL wins M4 Quarterly low and M4 Monthly low. Stable Gibbs also appears in the top three in several regimes. However, robust and performance-based baselines remain strongest in some cases. In particular, high disagreement M4 Quarterly favors the median and trimmed mean, while M4 Yearly low favors top-3 averaging and inverse-loss weighting.

This supports a careful interpretation. The results do not show that GGEW universally replaces classical forecast-combination methods. They show that online Gibbs-style adaptation can be competitive and sometimes leading, especially when the forecast pool contains exploitable disagreement structure.

\subsection{Monash Rolling-Origin Results}

The rolling-origin experiments are the most realistic online deployment tests in this study.

\begin{table}[ht]
\centering
\caption{Top Monash rolling-origin results.}
\scriptsize
\setlength{\tabcolsep}{3pt}
\renewcommand{\arraystretch}{1.02}
\begin{adjustbox}{max width=\textwidth}
\begin{tabular}{llrlr}
\toprule
Dataset & Rank & Method & Overall loss & Mean states evaluated \\
\midrule
Traffic Hourly & 1 & Local-UCB Stable Gibbs-NCL & 421.1985 & 4.05 \\
Traffic Hourly & 2 & Local-UCB Stable Gibbs & 421.2012 & 3.00 \\
Traffic Hourly & 3 & top-3 average & 423.2206 & -- \\
Traffic Hourly & 4 & Local-UCB Symmetric Gibbs & 423.2364 & 4.05 \\
Traffic Hourly & 5 & best historical model & 445.3267 & -- \\
\addlinespace[2pt]
Electricity Hourly & 1 & Local-UCB Stable Gibbs & 19,561,022,006 & 3.00 \\
Electricity Hourly & 2 & Local-UCB Symmetric Gibbs & 19,595,345,221 & 4.02 \\
Electricity Hourly & 3 & inverse-loss weighted & 19,629,569,576 & -- \\
Electricity Hourly & 4 & top-3 average & 19,706,911,054 & -- \\
Electricity Hourly & 5 & Local-UCB Stable Gibbs-NCL & 19,767,997,849 & 4.02 \\
\addlinespace[2pt]
Solar Weekly & 1 & Local-UCB Stable Gibbs & 17,890,727,069 & 3.00 \\
Solar Weekly & 2 & Local-UCB Symmetric Gibbs & 17,964,395,448 & 4.36 \\
Solar Weekly & 3 & Local-UCB Stable Gibbs-NCL & 18,005,974,186 & 4.36 \\
Solar Weekly & 4 & top-3 average & 18,219,076,776 & -- \\
Solar Weekly & 5 & inverse-loss weighted & 18,821,706,681 & -- \\
\bottomrule
\end{tabular}
\end{adjustbox}
\end{table}

``Mean states evaluated'' reports the average number of Local-UCB parameter states evaluated per forecasting case; it is not applicable to non-Gibbs baselines.

On Traffic Hourly, Local-UCB Stable Gibbs-NCL obtains the lowest aggregate loss, with Local-UCB Stable Gibbs almost tied. On Electricity Hourly and Solar Weekly, Local-UCB Stable Gibbs achieves the lowest aggregate loss, followed by Local-UCB Symmetric Gibbs. Solar Weekly is a smaller robustness experiment, but it is notable that the online Gibbs-family methods occupy the leading ranks.

\FloatBarrier

\subsection{Overall Empirical Pattern}

Across the main experiments, the empirical pattern is positive but not universal. Overall loss denotes the sum of squared forecast errors over all evaluated cases and forecast horizons.

First, online Gibbs-family methods are frequently competitive and sometimes leading. They win several M4 regimes and achieve the best aggregate losses in Traffic Hourly, Electricity Hourly, and Solar Weekly.

Second, classical baselines remain important. Median, trimmed mean, inverse-loss weighting, exponential weighting, and top-3 average are strong in several M4 settings. This confirms that the proposed method should be presented as an adaptive addition to the forecast-combination toolkit, not as a universal replacement for existing aggregation rules.

Third, disagreement structure appears important. In some high disagreement settings, robust aggregation remains strongest. In other high disagreement settings, such as M4 Monthly high, symmetric Gibbs performs best. This means the paper should avoid a simple rule such as ``high disagreement always favors robust methods.'' A more careful statement is that forecast disagreement structure affects which aggregation rule works best, and the proposed framework provides a way to study and adapt to that structure.

The main empirical conclusion is therefore:
\begin{quote}
\itshape
Adaptive forecast combination should be evaluated together with the structure of the forecast pool, the online deployment protocol, and the disagreement among base forecasters.
\end{quote}

\FloatBarrier

\section{Limitations}

This work has several limitations. First, the proposed method requires a pool of base forecasts. Its performance depends on the quality, diversity, and stability of that forecast pool. If all base models are poor or if the forecast pool lacks useful diversity, adaptive weighting cannot recover information that is not present in the candidate forecasts.

Second, the disagreement-regime stratification is empirical. Relative cross-model forecast variance is transparent and scale-adjusted, but it is not claimed to be the only possible measure of disagreement. Other measures, such as forecast entropy, pairwise correlation, rank disagreement, or residual covariance, may provide additional insight.

Third, the online Local-UCB adaptation currently operates over a discrete hyperparameter state space. This makes the method computationally practical and reproducible, but it does not solve continuous hyperparameter optimization. Continuous adaptation of the learning rate, diversity strength, or exploration parameter remains a useful extension.

Fourth, the Monash Traffic, Electricity, and Solar experiments use deterministic rolling baseline forecasts generated from raw public datasets. These experiments are useful for testing online deployment behavior, but they are not directly comparable to M4 public submission-matrix experiments, where the forecast pool comes from external submitted forecasting systems.

Finally, the framework focuses on point forecasts and squared-loss evaluation. Extensions to probabilistic forecasting, conformal calibration, quantile forecasting, and interval forecast combination remain important directions for future work.

\section{Conclusion}

\begingroup
\setlength{\parskip}{0.75em}
\setlength{\parindent}{0pt}

This paper introduced Generalized Gibbs Ensemble Weighting, a probabilistic framework for adaptive forecast combination. The method converts normalized predictive-loss scores into ensemble weights using a Gibbs-style exponential allocation rule, stabilizes the optimization through exponentiated-gradient updates, and extends the score with diversity-aware corrections.

The framework provides a single algorithmic recipe for Stable Gibbs, Directional Gibbs-NCL, and Symmetric Gibbs-NCL. It also supports online hyperparameter adaptation through Local-UCB state selection, which adapts the learning rate, diversity strength, and Gibbs variant without exhaustively evaluating the full hyperparameter grid at every prediction step.

Empirical evaluation across M4, Traffic, Electricity, and Solar experiments suggests that Gibbs-style adaptive weighting is a useful and competitive forecast-combination tool. The results do not support a universal dominance claim. Instead, they show that the usefulness of adaptive Gibbs weighting depends on the dataset, forecast horizon, forecast pool, deployment protocol, and disagreement structure.

The central contribution is therefore both methodological and empirical: GGEW provides a unified adaptive ensemble-weighting framework, and the experiments motivate a regime-aware view of forecast combination. Future work should study broader disagreement diagnostics, continuous online adaptation, probabilistic forecast extensions, and larger-scale deployment settings.

\endgroup

\section*{Author Contributions}

\begingroup
\setlength{\parskip}{0.6em}
\setlength{\parindent}{0pt}

P.R.N. conceived the forecasting framework, developed and implemented the algorithms, designed and executed the empirical evaluation, analyzed the results, and prepared the manuscript.

N.K.G. contributed to discussions connecting the proposed framework to Gibbs-style mathematical formulations and provided feedback on the mathematical derivations.

\endgroup

\end{document}